\documentclass{article}

\usepackage{amsmath,graphicx,subfigure,color}
\usepackage{multirow}
\usepackage{hyperref}
\usepackage[normalem]{ulem}
\usepackage{algorithm}
\usepackage{algpseudocode}
\usepackage[outercaption]{sidecap}    
\usepackage[listings]{tcolorbox}

\usepackage[nonatbib,final]{nips_2018}

\usepackage[utf8]{inputenc} % allow utf-8 input
\usepackage[T1]{fontenc}    % use 8-bit T1 fonts
\usepackage{hyperref}
\usepackage{cleveref}
\usepackage{url}            % simple URL typesetting
\usepackage{booktabs}       % professional-quality tables
\usepackage{amsfonts}       % blackboard math symbols
\usepackage{nicefrac}       % compact symbols for 1/2, etc.
\usepackage{microtype}      % microtypography

\title{A Study on Dialogue Reward Prediction for Open-Ended Conversational Agents}

\author{
  Heriberto Cuay\'ahuitl$^1$, Seonghan Ryu$^2$, Donghyeon Lee$^2$, Jihie Kim$^2$\\
  $^1$University of Lincoln, School of Computer Science, Lincoln, United Kingdom\\
  $^2$Samsung Research, Artificial Intelligence Team, Seoul, South Korea\\
  \texttt{HCuayahuitl@lincoln.ac.uk}, \texttt{\{seonghan.ryu,dh.semko.lee,jihie.kim\}@samsung.com} \\
}

\begin{document}
% \nipsfinalcopy is no longer used

\maketitle

\begin{abstract}
The amount of dialogue history to include in a conversational agent is often underestimated and/or set in an empirical and thus possibly naive way. This suggests that principled investigations into optimal context windows are urgently needed given that the amount of dialogue history and corresponding representations can play an important role in the overall performance of a conversational system. This paper studies the amount of history required by conversational agents for reliably predicting dialogue rewards. The task of dialogue reward prediction is chosen for investigating the effects of varying amounts of dialogue history and their impact on system performance. Experimental results using a dataset of 18K human-human dialogues report that lengthy  dialogue histories of at least 10 sentences are preferred (25 sentences being the best in our experiments) over short ones, and that lengthy histories are useful for training dialogue reward predictors with strong positive correlations between target dialogue rewards and predicted ones.
\end{abstract}

\section{Introduction}
Deep Reinforcement Learning holds a lot of promise for training intelligent conversational machines \cite{SimpleDS,Gao2018,Deng2018}, which can infer their behaviour from trial and error via interaction with an environment in order to optimise a long term reward signal \cite{SuttonB2018,Li17b}. This machine learning paradigm is interesting and worth studying because it resembles human learning \cite{schultz:science97,MnihKSRVBGRFOPB15}, and has wide application from low-level to high-level control using a single or multiple modalities. However, its successful application is not trivial due to interactions with real world environments (not easy to model), complex decision-making at different levels of granularity (not easy to infer or specify), and unkwown reward signals (also not easy to infer or specify), among others. In this paper we focus our attention to the latter only, the reward signals, in the context of conversational agents. Specifically, we focus on open-ended dialogue agents because they have received little attention so far in the literature---as described in Section~\ref{literature_review}. This type of agents are interesting and challenging at the same time because the amount of input features and their combinations are simply vast, making decision-making difficult. Even when some reward functions have been proposed by previous works \cite{LiMSJRJ16,LiMSJRJ17,SerbanEtAl2018long} it is unclear what is the best reward function to use for training future conversational agents or chatbots.

Our study uses a dataset of human-human dialogues, which we extend with noisy dialogues by replacing human responses with randomly chosen responses from other dialogues. While non-noisy dialogues aim to show human-like and desirable outputs, the noisy ones aim to exemplify less desirable behaviour. Each dialogue is automatically labelled with a numerical reward according to the amount of noise (distorted sentences)---the automatic labels facilitate its application to other datasets. This extended dataset is used for predicting dialogue rewards of open-ended conversations. As part of our study, we propose a new reward function that is easy to implement and that strongly correlates with test human-human dialogues. The latter is indeed possible by using lengthy dialogue histories. 

%Motivations:
%(1) Unknown size of dialogue history in chit-chat systems
%(2) Automatically labelled data of open-ended dialogues
%(3) Reliable prediction of human-likeness in generated dialogues
%(4) Previous related works focus on task-oriented dialogues

\section{Related Work}
\label{literature_review}
So far little attention has been devoted to predicting numerical rewards of open-ended dialogues, which can be used to optimise the behaviour of dialogue agents such as chit-chat chatbots. By dialogue reward prediction we mean `estimating the quality of human-computer or human-human dialogues', where quality can be measured for example via task success, user satisfaction or human-likeness, among others. Most previous related works focus on task-oriented dialogue systems, which require labelled data typically collected by a previously developed dialogue system. For example, \cite{SuGY18} used previously collected human-computer dialogues in the domain of restaurants for training an LSTM-based dialogue regressor for predicting rewards of the form $\mathcal{R}=20 \times 1_{success} -N$, where a value of 20 is given for a successful dialogue minus the number of dialogue turns. Such a regressor---which can be turned into a binary classifier for predicting success or failure---was re-trained online together with a dialogue policy using active learning. \cite{NoseworthyCP17} also use LSTM-based predictors of dialogue success (or failure) in the programming domain (StackOverflow data) based on word-based features, where success is given to a correctly answered question---failure otherwise. Similarly,  \cite{PapangelisKS17} explores other machine learning techniques for predicting dialogue success but from acoustic features instead of word-based features. \cite{UltesBCMRSWGY17} predict dialogue rewards of the form $\mathcal{R}=(iq-1) \times 5 -N$, where $iq$ is an interaction quality value derived from a multiclass classifier. While speech recognition features and dialogue actions are used as inputs, the outputs used for training were obtained from three human labellers that judged the quality of the interactions. The reward predictor induced from a dataset of bus information was used for training policies in the domains of restaurants, hotels and laptops. 

The related studies above, motivated by the earlier work of \cite{WalkerLHWG02} in predicting problematic dialogues, have focused on task-oriented dialogues that aim to achieve a goal such as finding restaurant/hotel/laptop information, call routing, etc., see summary in Table~\ref{litreview}. There is however little related research on human-human or human-computer dialogues where there is no clear notion of a task or goal to achieve---notable exceptions are \cite{SerbanEtAl2018long,LiMSJRJ17,LiMSJRJ16}. \cite{SerbanEtAl2018long} train a neural multiclass classifier and linear regressor from human-chatbot dialogues from Amazon Mechanical Turk (AMT). In this study, AMT human evaluators are given dialogues and are asked to rate the quality of candidate responses according to the following categories (or labels): 1 indicates an innapropriate response or one that does not make sense, 3 indicates neutral, and 5 is a highly appropriate and interesting response. The classifier receives inputs with 1458 features (including lexical, syntactic and semantic information) and outputs a label out of the five possible labels. The regressor aims to learn a reward function for open-ended dialogues. It receives inputs with 23 features (also including lexical, syntactic and semantic information) and outputs a real value between 1 and 5. This regressor was used to train neural-based reinforcement learning dialogue policies for open-ended dialogues -- \cite{MisuGLT12} used a similar approach in a restricted set of domains. \cite{LiMSJRJ17} train a binary classifier for predicting whether a dialogue is human-generated or machine generated. The outputs of this classifier, derived from a two-class Softmax function, are used as dialogue rewards in order to train a generator that aims for deceiving such a classifier within an adversarial framework. \cite{LiMSJRJ16} produces rewards of the form $r(a,[p_i,q_i]) =\lambda_1 r_1 + \lambda_2 r_2 + \lambda_3 r_3$, where $a$ refers to the generated response, $[p_i,q_i]$ are the previous two dialogue turns, $r_1$ and $r_3$ refer to likelihood outputs of $Seq2Seq$ models, $r_2$ refers to dialogue history similarity, and $\lambda_i$ are empirically defined weights (easy of answering: $\lambda_1=0.25$, information flow: $\lambda_2=0.25$, and semantic coherence: $\lambda_3=0.5$). While \cite{SerbanEtAl2018long} reports a 0.40 Pearson correlation coefficient between predicted rewards and AMT human judgements, it is unclear whether or not the binary classifier proposed by \cite{LiMSJRJ17} or heuristic function of \cite{LiMSJRJ16} can produce rewards that correlate with human judgements.
%Other related works train similar dialogue policies by using predefined rewards.
%(called `Supervised AMT')
%(called `Reward Model')

The work presented in the remaining of the paper goes beyond previous works as follows: 
\begin{enumerate}
\item we study dialogue reward prediction for open-ended dialogues using automatically labelled data. This is in contrast with \cite{SerbanEtAl2018long,MisuGLT12} who use paid human evaluators,  which results in an expensive and time consuming process because large amounts of dialogues require labelling; 
\item we use contextually-richer and automatically derived input features to represent dialogue histories. This is in contrast with \cite{SerbanEtAl2018long,MisuGLT12} who use a small set of carefully designed features and with \cite{LiMSJRJ16,LiMSJRJ17,SerbanEtAl2018long} who use short dialogue histories for dialogue reward prediction; and 
\item we investigate the impact of dialogue history length in dialogue reward prediction for open-ended conversational agents. Our best predictor---using rich dialogue histories---reports a strong positive correlation on unseen test dialogues, which is higher than the previously reported scores of \cite{SerbanEtAl2018long} with only moderate positive correlation.
\end{enumerate}

\begin{table}[h!]
  \begin{center}
    \begin{tabular}{|l|l|l|}
\hline
Authors & Domain(s) & Model\\
\hline
\hline
Su et al. \cite{SuGY18} & Restaurants & Recurrent Neural Network (LSTM)\\
\hline
Noseworthy et al. \cite{NoseworthyCP17} & Programming (StackOverflow) & Flat and Hierarchical LSTMs\\
\hline
Papangelis et al. \cite{PapangelisKS17} & Laptops, Restaurants & Support Vector Machines, Gaussian \\
 & & Process Regressors, Random Forests\\
\hline
Ultes et al. \cite{UltesBCMRSWGY17} & Bus, Restaurants, Hotels, Laptops  & Support Vector Machine (SVM)\\
\hline
Walker et al. \cite{WalkerLHWG02} & Call Routing & Rule Learner (RIPPER)\\
\hline
Misu, et al. \cite{MisuGLT12} & Restricted QA in a  Museum & Unknown\\
\hline
Li et al. \cite{LiMSJRJ16,LiMSJRJ17} & Open-Ended & Recurrent Neural Network (LSTM)\\
\hline
Serban at al. \cite{SerbanEtAl2018long} & Open-Ended & Fully-Connected Neural Networks\\
\hline
    \end{tabular}
  \caption{Related works on dialogue reward or dialogue success prediction}
  \label{litreview}
  \end{center}
\end{table}

%\newpage
\section{Proposed Study}
\label{study}
%\begin{tcolorbox}
We aim to answer the research questions {\bf How dialogue rewards can be predicted in open-ended conversations? How much dialogue history do open-ended conversational agents need?} 
%\end{tcolorbox}

This study focuses on open-ended dialogues as those observed in chit-chat interactions. On the one hand, the longer the history the richer the context to take into account, but the more computational expense involved (due to high amounts of features). On the other hand, the shorter the history the simpler the dialogue context at a cheaper computational expense. In addition, we study the effects of short and long dialogue histories in the task of dialogue reward prediction---often used for dialogue optimisation---and we treat dialogue reward prediction as a regression problem. 

Let $\mathcal{D}=\{d_1,\dots,d_N\}$ be a set of open-ended human-human conversations in raw text, where each dialogue $d_i=\{t_1,\dots,t_T\}$ contains a set of dialogue turns $t^i_j=\{s^1_1,\dots,s^1_X\}$, and each sentence $s^i_{j,k}=\{w^1_{1,1},\dots,w^1_{1,Y}\}$ contains a set of words as part of the verbal contribution in dialogue turn $i,j$. We use $T=|d_i|$ to denote the length---in terms of number of dialogue turns---of dialogue $i$. Assuming two partner conversants in each dialogue (our case), a dialogue turn has sentences of the form $s^i_j=\{(s^i_{j_A},s^i_{j_B})\}$---one per conversant in each dialogue turn, e.g. human A says sentence $s^i_{1_A}$, human B says sentence $s^i_{1_B}$, then human A says sentence $s^i_{2_A}$, human B says sentence $s^i_{2_B}$, and so on until the end of  dialogue $i$.  

We propose to derive dialogue rewards with automatically labelled data (for practical application) by extending a dataset of human-human dialogues and assigning {\it positive values} to responses seen in dataset $\mathcal{D}$, and {\it negative values} to randomly generated responses due to lacking coherence (also referred to as `non-human-like responses'). A complete dialogue can be rewarded as 
\vskip-5pt
\begin{equation}\nonumber
R_i=\sum_{j=1}^T r_j^i(a),
\end{equation}
where $i$ is the dialogue in focus, $j$ the dialogue turn in focus, and $r_j^i(a)$ is given according to
\vskip-5pt
\begin{equation}\nonumber
  r^i_j(a)=\begin{cases}
    +1, & \text{if $a$ is a human response in dialogue turn $(i,j)$}.\\
    -1, & \text{if $a$ is human but randomly chosen (incoherent)}.
  \end{cases}
\end{equation}

Table~\ref{exampledialogue_good} shows an example of a well rewarded dialogue (without distortions) and Table~\ref{exampledialogue_bad} shows an example of a poorly rewarded dialogue (with distortions). Other dialogues can exhibit similar dialogue rewards or something in between (ranging between $-T$ and $T$), depending on the amount of distortions---the higher the amount of distortions the lower the dialogue reward. 

\begin{table}[bh!]
  \begin{center}
    \begin{tabular}{|c|l|c|}
\hline
turn & Verbalisation & Reward\\
\hline
\multirow{2}{*}{1} & A: hello what are doing today ? &\\
 & B: i'm good , i just got off work and tired , i have two jobs . & +1\\
\hline
\multirow{2}{*}{2} & A: i just got done watching a horror movie&\\
 & B: i rather read , i have read about 20 books this year . & +1\\
\hline
\multirow{2}{*}{3} & A: wow ! i do love a good horror movie . loving this cooler weather&\\
 & B: but a good movie is always good .& +1\\
\hline
\multirow{2}{*}{4} & A: yes ! my son is in junior high and i just started letting him watch them too&\\
 & B: i work in the movies as well .& +1\\
\hline
\multirow{2}{*}{5} & A: neat ! ! i used to work in the human services field&\\
 & B: yes it is neat , i stunt double , it is so much fun and hard work . & +1\\
\hline
\multirow{2}{*}{6} & A: yes i bet you can get hurt . my wife works and i stay at home&\\
 & B: nice , i only have one parent so now i help out my mom . & +1\\
\hline
\multirow{2}{*}{7} & A: i bet she appreciates that very much .&\\
 & B: she raised me right , i'm just like her . & +1\\
\hline
\multirow{2}{*}{8} & A: my dad was always busy working at home depot&\\
 & B: now that i am older home depot is my toy r us .& +1\\
\hline
    \end{tabular}
  \caption{Example human-human dialogue with a good numerical reward $\sum_{j=1}^{T=8} r_j^1(a)=8$}
  \label{exampledialogue_good}
  \end{center}
\end{table}

We employ Algorithm~\ref{dgen} for generating dialogues with varying amounts of distortions (i.e. different degrees of human-likeness), which will be used for training and testing reward prediction models. Given our extended dataset $\mathcal{\hat{D}}=\{(\hat{d}_1,y_1),\dots,(\hat{d}_N,y_N)\}$ with (noisy) dialogue histories $\hat{d}_i$, the goal is to predict dialogue scores $y_i$ as accurately as possible.

\begin{table}[t!]
  \begin{center}
    \begin{tabular}{|c|l|c|}
\hline
Turn & Verbalisation & Reward\\
\hline
\multirow{3}{*}{1} & A: hello what are doing today ? &\\
 & B: \sout{i'm good , i just got off work and tired , i have two jobs .} & \sout{+1}\\
 & B: lol. she is more pop now. her old stuff was country. i like classic country . & -1\\
\hline
\multirow{3}{*}{2} & A: i just got done watching a horror movie&\\
 & B: \sout{i rather read , i have read about 20 books this year .} & \sout{+1}\\
 & B: i am good how about you & -1\\
\hline
\multirow{3}{*}{3} & A: wow ! i do love a good horror movie . loving this cooler weather&\\
 & B: \sout{but a good movie is always good .}& \sout{+1}\\
 & B: i do not enjoy the driving in my honda civic & -1\\
\hline
\multirow{3}{*}{4} & A: yes ! my son is in junior high and i just started letting him watch them too&\\
 & B: \sout{i work in the movies as well .}& \sout{+1}\\
 & B: do you like dogs ? I have two .& -1\\
\hline
\multirow{3}{*}{5} & A: neat ! ! i used to work in the human services field&\\
 & B: \sout{yes it is neat , i stunt double , it is so much fun and hard work .} & \sout{+1}\\
 & B: very cool . i had one of those . & -1\\
\hline
\multirow{3}{*}{6} & A: yes i bet you can get hurt . my wife works and i stay at home&\\
 & B: \sout{nice , i only have one parent so now i help out my mom .} & \sout{+1}\\
 & B: that is just great . we need more educators like you in the system . & -1\\
\hline
\multirow{3}{*}{7} & A: i bet she appreciates that very much .&\\
 & B: \sout{she raised me right , i'm just like her .} & \sout{+1}\\
 & B: is it bad that i hate pumpkins and trees ? & -1\\
\hline
\multirow{3}{*}{8} & A: my dad was always busy working at home depot&\\
 & B: \sout{now that i am older home depot is my toy r us .}& \sout{+1}\\
 & B: why not ? maybe you do not like to travel ? & -1\\
\hline
    \end{tabular}
  \caption{Example distorted human-human dialogue with a poor numerical reward $\sum_{j=1}^8 r_j^2(a)=-8$}
  \label{exampledialogue_bad}
  \end{center}
\end{table}

\begin{algorithm} [h!]
\footnotesize
\caption{\label{ChatDQN} Generator of (Distorted) Dialogues}\label{dgen} 
\begin{algorithmic}[1]
\State $\mathcal{\hat{D}} \leftarrow [\emptyset]$ \Comment Initialise dataset of original and distorted dialogues
\For{dialoge $i=1,\dots,N$ in set $\mathcal{D}$} \Comment Iterate over the set of original dialogues
  \For{noise $n=0,\dots,|d_i|$}  \Comment Iterate over different amounts of noise ($|d_i|=$ no. dialogue turns)
    \State $score=|d_i|-(n*2)$ \Comment Score range: $(|d_i|\times-1),\dots,|d_i|$
    \State $\hat{d} \leftarrow [\emptyset]$ \Comment Initialise distorted dialogue
    \For{dialogue turn $j=0,\dots,T-1$ in dialogue $i$} \Comment Iterate over all sentences in dialogue $i$
      \State $s^i_j \leftarrow$ sentence of first partner conversant
      \State $s^i_{j+1} \leftarrow$ sentence of second partner conversant
      \If{$j<n$}
        \State $\hat{s}^i_{j+1} \leftarrow$ distorted sentence of second partner conversant
        \State $\hat{d} \leftarrow $ append tuple $<s^i_j,\hat{s}^i_{j+1}>$ \Comment Append distorted dialogue turn
      \Else
        \State $\hat{d} \leftarrow $ append tuple $<s^i_j,s^i_{j+1}>$ \Comment Append non-distorted dialogue turn
      \EndIf
      \State $j=j+1$
    \EndFor
    \State $\mathcal{\hat{D}} \leftarrow $ append tuple $<\hat{d}, score>$ \Comment Append newly generated (noisy) scored dialogue 
    \State $n=n+1$
  \EndFor
  \State $i=i+1$
\EndFor
\end{algorithmic}
\end{algorithm}

\section{Data}
We used data from the {\it Persona-Chat} dataset\footnote{Dataset downloaded from \url{http://parl.ai/} on 18 May 2018 \cite{MillerFBBFLPW17}}, which includes 17,877 dialogues for training and 999 dialogues for testing. The average dialogue length is 7.35 and 7.8 dialogue turns (166 and 186 words per dialogue, including punctuation) for training and testing, respectively. The vocabulary size in the entire dataset contains 19,667 unique words. See example dialogue in Table~\ref{exampledialogue_good}. We used this dataset as seed dialogues in order to generate dialogues with different amounts of noise (or distortions)---as described in the previous section---and obtained 149,308 dialogues for training and 8,704 for testing.

\section{Experiments and Results}

\subsection{Experimental Setting}
We represent a dialogue history via its mean word vectors as in  Deep Averaging Networks \cite{IyyerMBD15}, where sentences are represented with numerical feature vectors denoted as ${\bf x}=\{x_1,...,x_{|{\bf x}|}\}$. In this way, a set of word sequences $s^i_j$ in dialogue-sentence pair $i,j$ is mapped to feature vectors 
\begin{equation}\nonumber
{\bf x}^i_j=\frac{1}{N^i_j}\sum_{k=1}^{N^i_j} c^i_{j,k},
\end{equation} 
where $c^i_{j,k}$ is the vector of coefficients of word $k$, part of sentence $j$ in dialogue $i$, and $N^i_j$ is the number of words in the sentence in focus.

To complete the preparation of our training and test data, vector ${\bf Y}=\{y_1,...,y_{|{\bf Y}|}\}$ is the set of target labels---generated as described in Section~\ref{study}. In this way, dataset $\mathcal{D}^{train}=({\bf X}^{train},{\bf Y}^{train})$ is used for training neural regression models using varying amounts of dialogue history, and dataset $\mathcal{D}^{test}=({\bf X}^{test},{\bf Y}^{test})$ is used for testing the learnt models.

\begin{figure}[t]
\centering
\includegraphics[width=85mm]{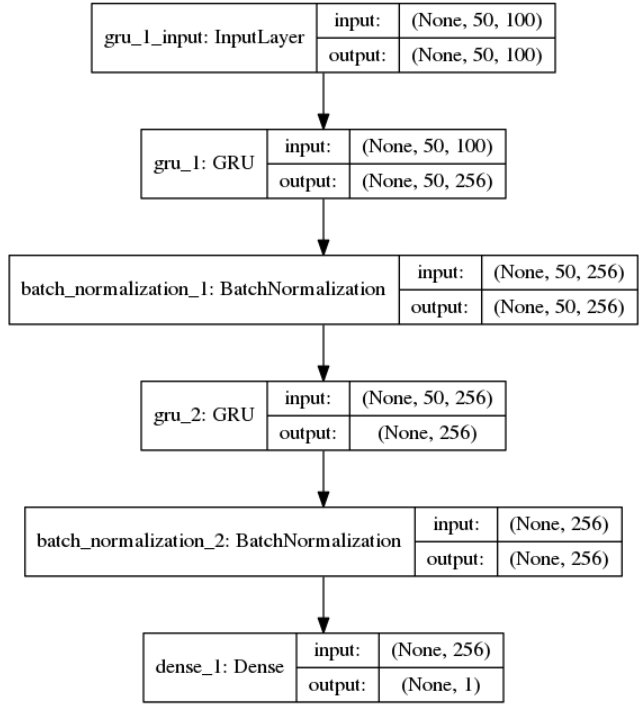}
\caption{Illustration of our recurrent neural network for dialogue reward prediction}
\label{RNN}
\end{figure}

All our experiments use a 2-layer Gated Recurrent Unit (GRU) neural network \cite{choEtAlEMNLP2014}, see Figure~\ref{RNN}. In this recurrent neural network---at each time step $t$ in a dialogue history---the first hidden layer generates a hidden state ${\bf h}_t$ as follows: 
\begin{equation}\nonumber
\begin{gathered}
{\bf r}_t=\sigma ({\bf W}_r {\bf x}_t + {\bf U}_r {\bf h}_{t-1}),\\
{\bf z}_t=\sigma ({\bf W}_z {\bf x}_t + {\bf U}_z {\bf h}_{t-1}),\\
\bar{\bf h}_t=\mbox{tanh} ({\bf W}_{\bar{\bf h}} {\bf x}_t + {\bf U}_{\bar{\bf h}} ({\bf r}_{t-1} \odot {\bf h}_{t-1})),\\
{\bf h}_{t}=\mbox{BN}_{\gamma,\beta} \left[ (1-{\bf z}_{t}) \odot {\bf h}_{t-1} + {\bf z}_{t} \odot \bar{{\bf h}_{t}} \right],
\end{gathered}
\end{equation}
%{\bf h}_{t}=\mbox{BN}_{\gamma,\beta}({\bf h}_{t}),
%{\bf o}_{t}=\mbox{BN}_{\gamma,\beta}({\bf W}_{o} {\bf h}_{t}),
where ${\bf r}_t$ is a reset gate that decides how much to forget the previous state, ${\bf z}_t$ is an update gate that decides how much it updates its activation, $\bar{\bf h}_t$ is an internal state, $\sigma(.)$ and $tahn(.)$ are the Sigmoid and hyperbolic Tangent functions (respectively), ${\bf W}_{*}$ and ${\bf U}_{*}$ are learnt weights, $\odot$ refers to the element-wise multiplication, and $\mbox{BN}_{\gamma,\beta}$ refers to Batch Normalisation with learnt weights $\gamma$ and $\beta$ \cite{IoffeS15}. Assuming that the equations above can be summarised as ${\bf h}_{t}=GRU({\bf x}_t,{\bf h}_{t-1})$ we get the following output taking into account both hidden layers $i$ in our neural network
\begin{equation}\nonumber
\begin{gathered}
{\bf h}_{t}^1=GRU({\bf x}_t,{\bf h}_{t-1}^1),\\
{\bf h}_{t}^2=GRU({\bf h}_{t}^1,{\bf h}_{t-1}^2),\\
o_{t}={\bf W}_{o} {\bf h}_{t}^2,
\end{gathered}
\end{equation}
where $o_{t}$ refers to the predicted dialogue reward for  dialogue history ${\bf x}_t$.
% using learnt representations ${\bf h}_{t}^1$ and ${\bf h}_{t}^2$.

We trained this neural network\footnote{Our experiments ran on a cluster of GPUs Tesla K80, and their implementation used the Keras library \url{https://github.com/keras-team/keras}} for varying amounts of dialogue lengths, where ${\bf x}_t$ included 1, 3, 5, 10, 25 and 50 sentences. Each sentence was represented with a mean word vector using the pretrained coefficients of the Glove model \cite{PenningtonSM14} (file glove.840B.300d.txt to be precise). Our experiments used the first 100 dimensions per word vector---empirical evidence reported that this dimensionality is a good compromise between prediction performance and computational expense. Note that the larger the word vector dimensionality the more parameters in the neural network. Other hyperparameters in our experiments include  batch size=128, dropout=0.2, dimensionality=256, loss function=Mean Absolute Error (MAE), optimiser=Adam, epochs=100, and early stopping=10 epochs. In addition, the training set was split into 80\% for training and 20\% for validation. The number of parameters in our neural network corresponds to 668K.

\subsection{Experimental Results}
\begin{figure}[t]
\centering
\includegraphics[width=90mm]{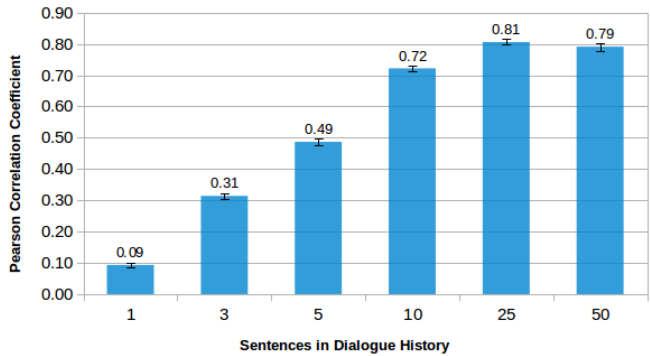}
\caption{Bar plot showing the performance of our dialogue reward predictors using different amounts of dialogue history (from 1 sentence to 50 sentences). Each bar reports an average Pearson correlation score over 10 runs, where the coefficients report the correlation between true dialogue rewards and predicted dialogue rewards in our test data}
\label{barplot}
\end{figure}

We trained neural networks for six different lengths of dialogue history, ranging from 1 sentence to 50 sentences. Each length size involved a separate neural network, trained 10 times in order to report results over multiple runs. Figure~\ref{barplot} reports the average Pearson correlation coefficient---between true dialogue rewards and predicted dialogue rewards---for each length size. It can be observed that short dialogue histories contribute to obtain weak correlations, and that longer dialogue histories ($\geq 10$ sentences) contribute to obtain strong correlations. It can also be observed that the longest history may not be the best choice of length size, the network using $25$ sentences achieved the best results.

%\begin{figure}[h!]
%\centering
%%\includegraphics[width=80mm]{pics/aaai-human-eval.eps}
%\includegraphics[width=80mm]{pics/dialhist-barplot.eps}
%\caption{Bar plot of six dialogue reward predictors using different amounts of dialogue history (from 1 sentence to 50 sentences) --- each bar reports an average Pearson correlation score over 10 runs}
%\label{barplot}
%\end{figure}

\begin{figure*}[t!]
\begin{center}
%\vspace{0.5cm}
\subfigure[1 Sentence ($r=0.09$)]{
\includegraphics[width=69mm]{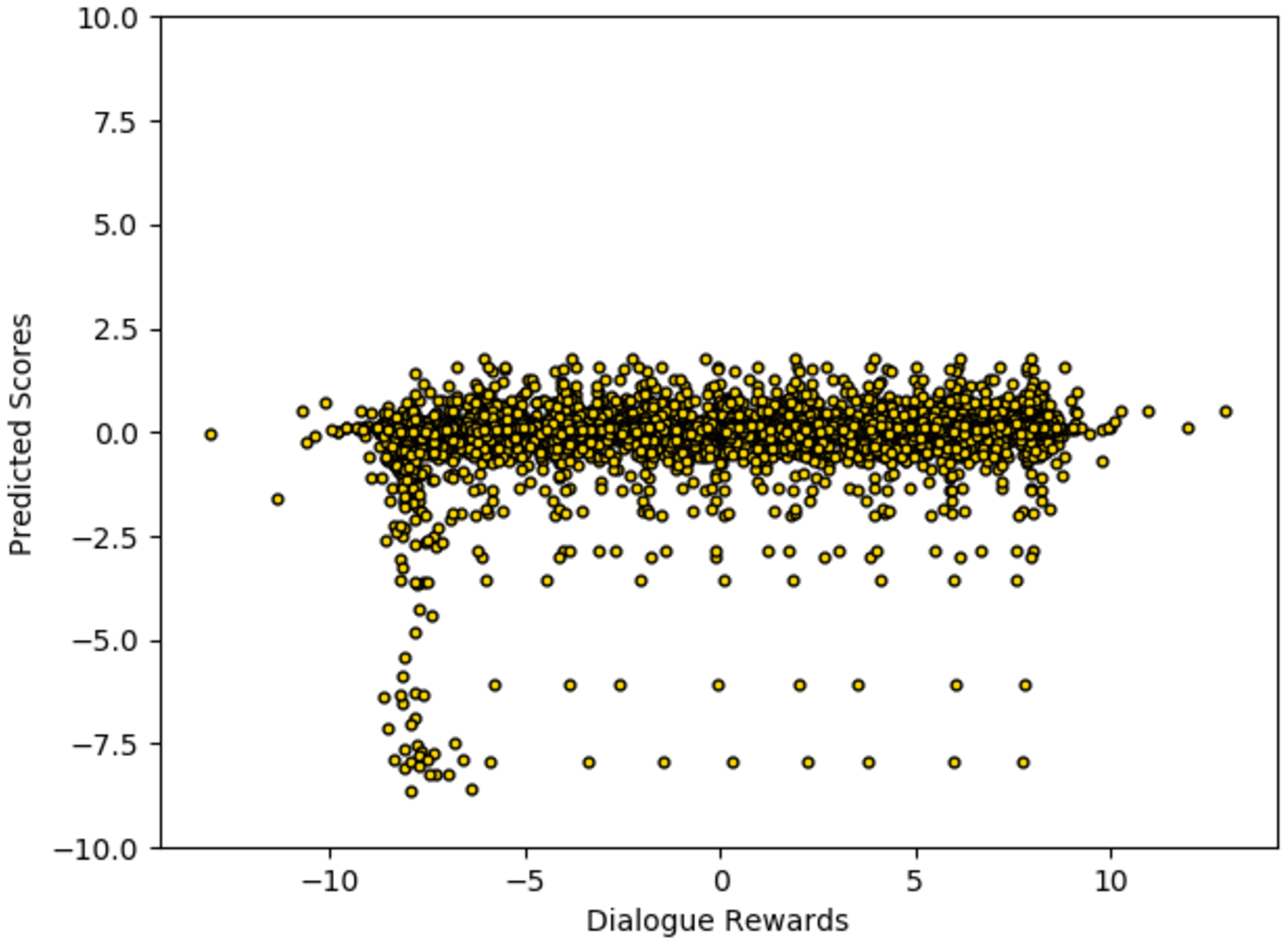}}
\hspace{-0.3cm}
\subfigure[3 Sentences ($r=0.31$)]{
\includegraphics[width=69mm]{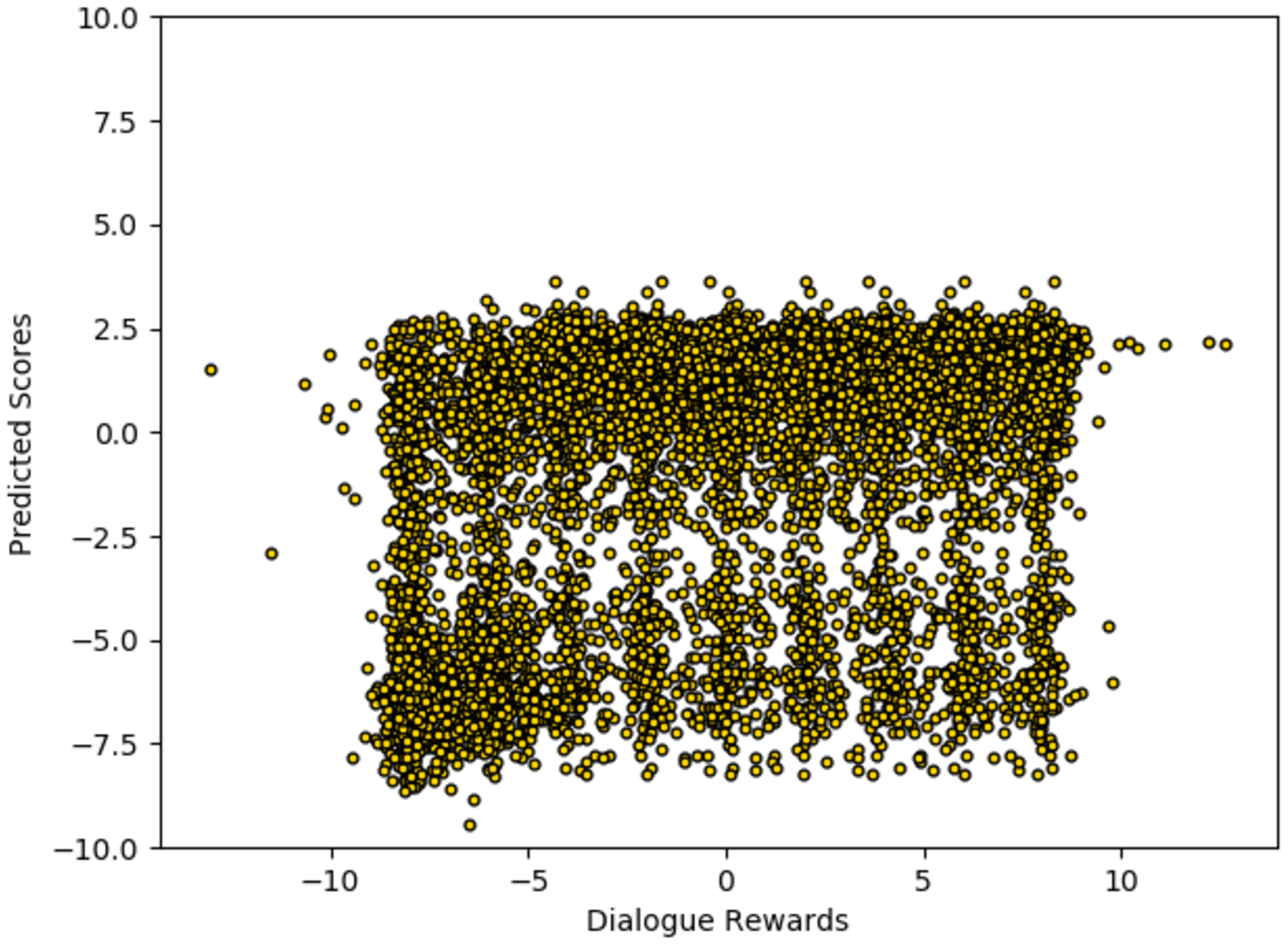}}
\hspace{-0.3cm}
\subfigure[5 Sentences ($r=0.49$)]{
\includegraphics[width=69mm]{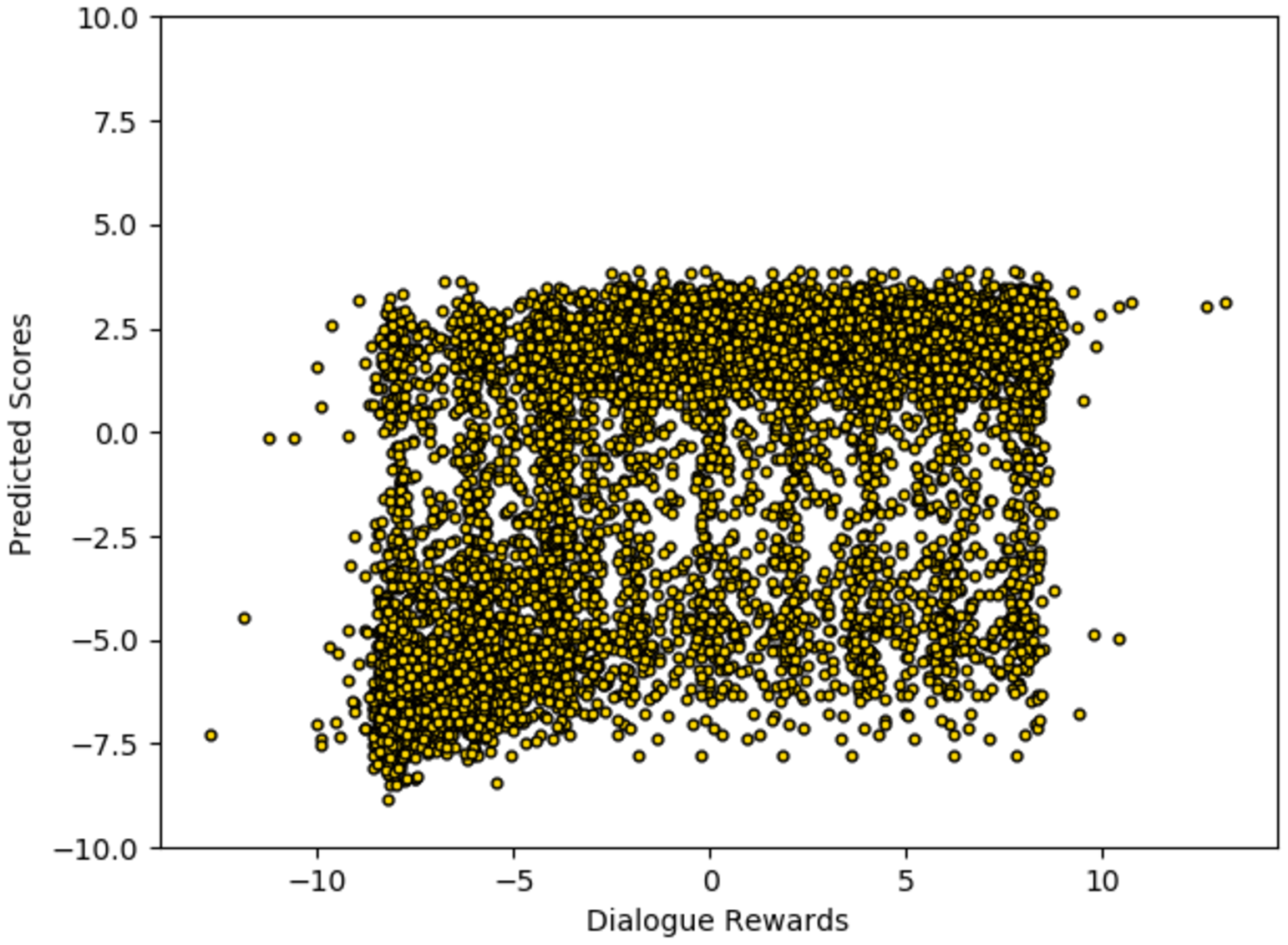}}
\hspace{-0.3cm}
\subfigure[10 Sentences ($r=0.72$)]{
\includegraphics[width=69mm]{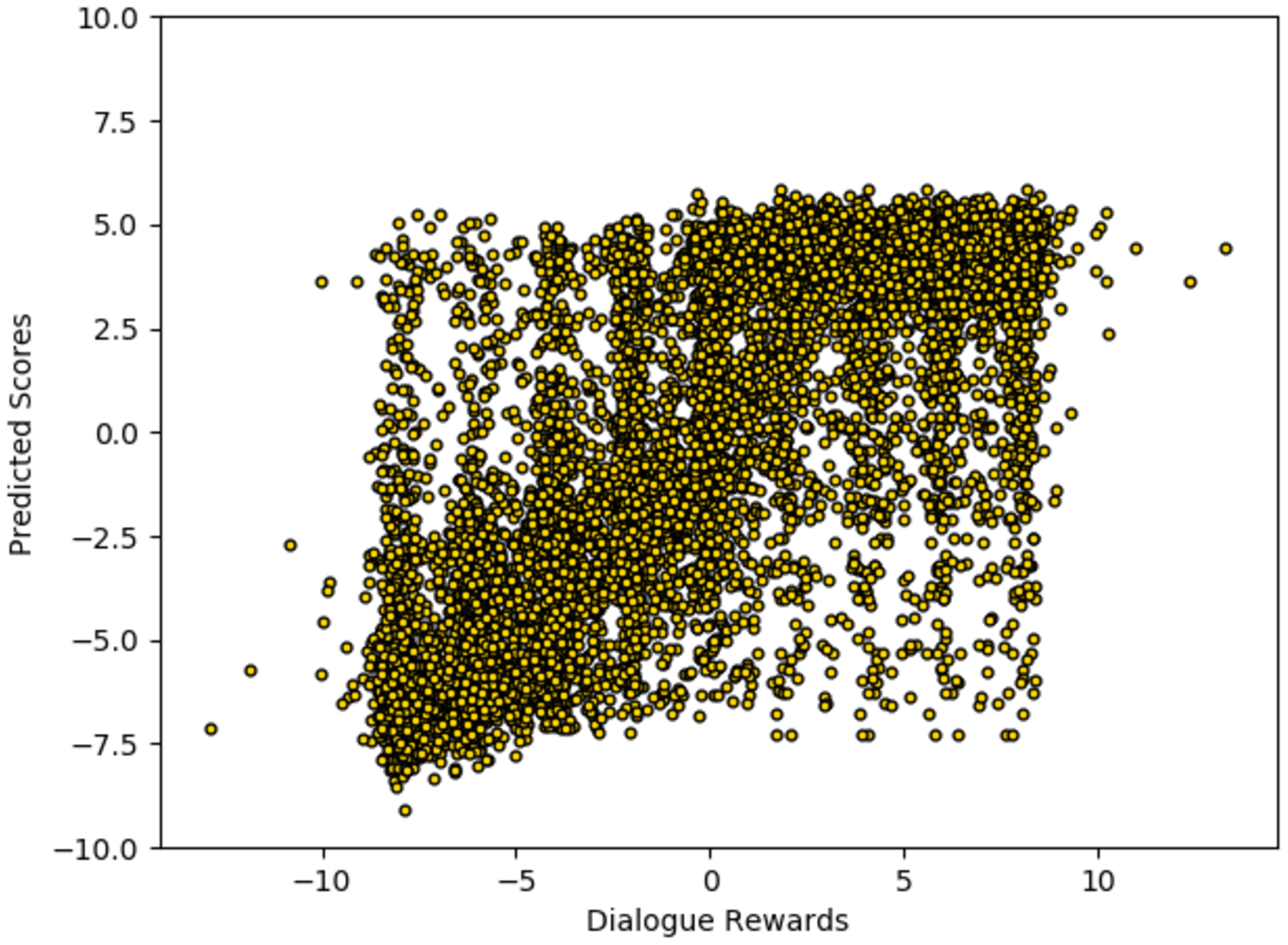}}
\hspace{-0.3cm}
\subfigure[25 Sentences ($r=0.81$)]{
\includegraphics[width=69mm]{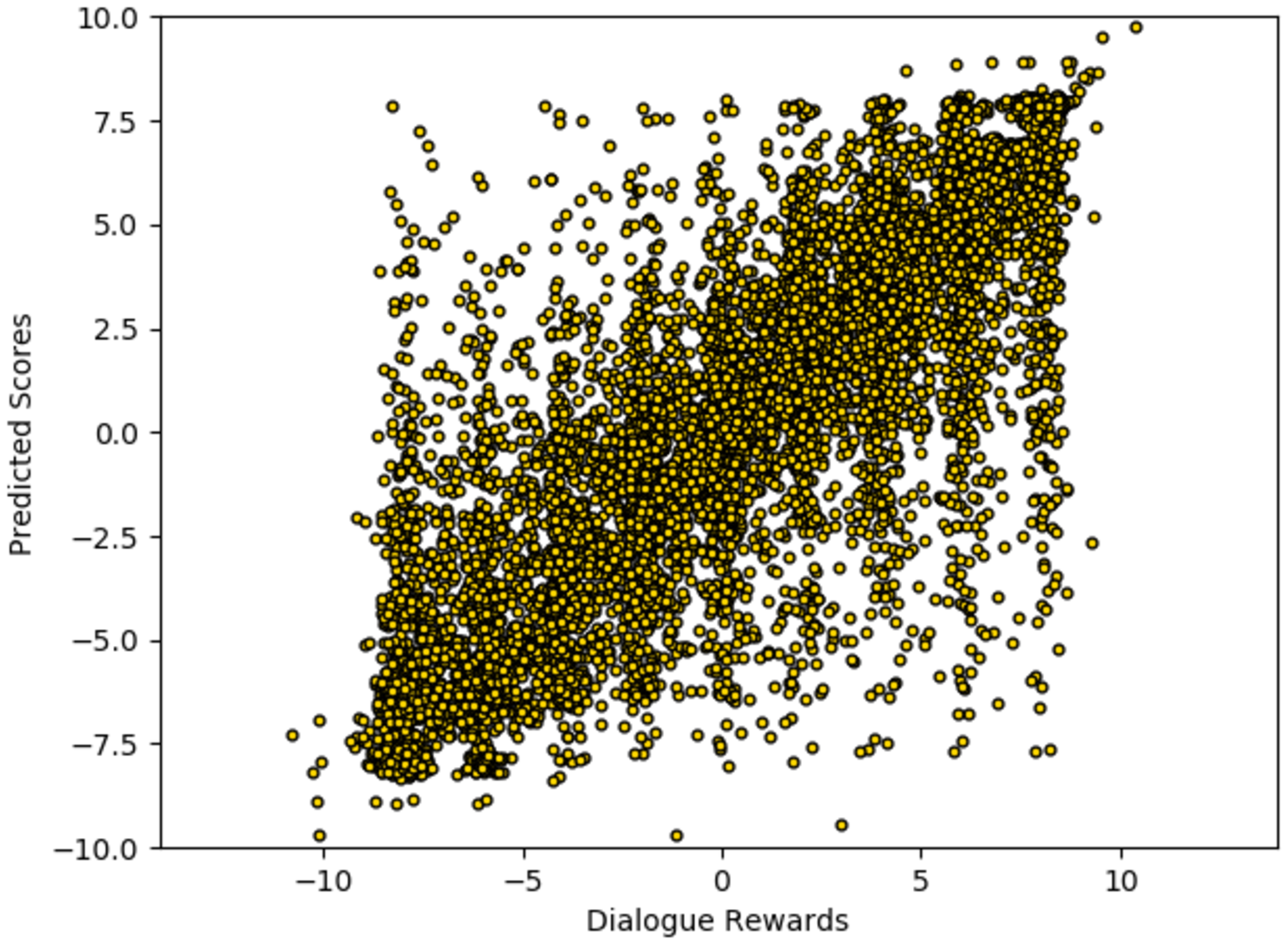}}
\hspace{-0.3cm}
\subfigure[50 Sentences ($r=0.79$)]{
\includegraphics[width=69mm]{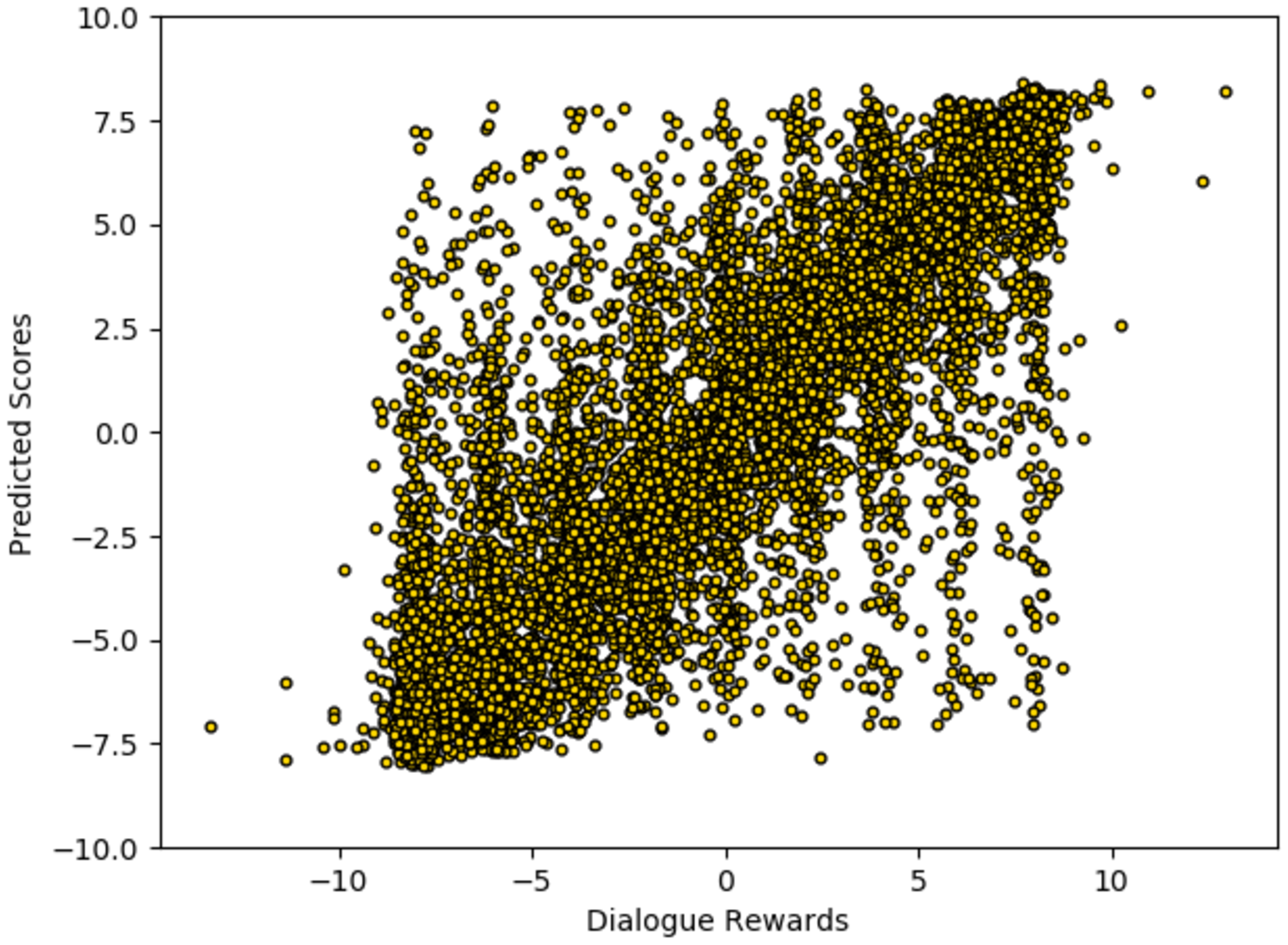}}
%}
\end{center}
\caption{Scatter plots of dialogue reward predictors using different amounts of sentences in dialogue history, where $r$=correlation coefficients and X-axis includes Gaussian noise drawn from $\mathcal{N}(0, 0.3)$}
%-- Figure~\ref{barplot} shows their corresponding Pearson correlation scores}
\label{scatterplots}
\end{figure*}

Figure~\ref{scatterplots} shows a more detailed inspection of our results using a scatter plot for the best trained neural networks, one per dialogue length size, chosen according to their performance in the validation set. These scatter plots show the correlation between target dialogue rewards (in our test dataset) and predicted dialogue rewards. The data points are test examples, where the number of data points is the same as the number of examples in our test dataset. The scatter plots include Gaussian noise drawn from $\mathcal{N}(0, 0.3)$ in the X-axis for better visualisation at the expense of showing less correlations.  It can be noted that 1 and 3 sentences in the dialogue history lead to weak correlations, 5 sentences in the dialogue history lead to moderate correlations, and 10 or more lead to strong positive correlations. From these results we can conclude that long (or very informative) dialogue histories should be used in open-ended conversational agents. 

\section{Conclusion and Future Work}
This paper studies dialogue reward prediction for open-ended conversational systems. Our study aimed at finding a reliable dialogue reward predictor---useful for dialogue optimisation---using data specified in a principled and practical way. To achieve this we propose a new reward function, train neural regressors from automatically labelled data, and evaluate their impact using different dialogue history lengths. 
%The goal is to find out what size(s) of dialogue history is/are recommendable for equipping conversational agents with amounts of dialogue context specified in a principled way. 
%We investigate such an impact in the task of dialogue reward prediction, useful for optimising dialogue agents. 
We adopted a dataset of human-human conversations and expanded it with noisy conversations in order to generate dialogues with varying amounts of noise, resulting in different dialogue rewards -- the higher the numerical reward the more human-like dialogue and the lower the numerical reward the less human-like dialogue. Our new noisy dataset was used to train neural regressors that took into account different amounts of dialogue history in order to observe the predictive power of target dialogue rewards. We modelled sentences in dialogue histories via their mean word vectors, using pre-trained word embeddings. Experimental results show that dialogue histories including 10 or more sentences lead to good performance (i.e. strong correlations between target dialogue rewards and predicted ones), and that shorter histories lead to poorer prediction power (i.e. moderate or weak correlations between target dialogue rewards and predicted ones). A history size of 25 sentences reported the best results in our test data of chit-chat conversations. Our findings are in contrast with related work that have confined themselves to short dialogue histories of 2 turns \cite{LiMSJRJ16} or 2 sentences \cite{LiMSJRJ17,SerbanEtAl2018long}.

Future conversational agents or chatbots can optimise their behaviour using neural regressors with lengthy or informative  dialogue histories.  Other future works can also investigate the impact of different dialogue history sizes in other datasets and types of dialogues (e.g. task oriented conversations).

\bibliographystyle{abbrv}
\bibliography{refs}

%This project aims to develop methodologies and algorithms for training a Chatbot humanoid robot. A platform-independent robot system will be developed with support for playing memory card games with embedded chit-chat. The targeted Chatty robot system aims for mimicking verbal behaviour aligned with body movements including eye blinking, head noddings, and arm movements such as pointing and gestures. The following objectives will be pursued.

%(1) visual perception

%(2) sentence+movement generation

%(3) coordination of the overall interaction

\end{document}